\documentclass[runningheads]{llncs}
\usepackage{graphicx}
\usepackage{xcolor}
\usepackage{tabularx}
\usepackage{multirow}
\usepackage{hyperref}
\usepackage{makecell}
\usepackage{algorithm}
\usepackage{algpseudocode}
\begin{document}
\title{Learning-Based Strategy Design for Robot-Assisted Reminiscence Therapy Based on a Developed Model for People with Dementia
}
\titlerunning{Learning-Based Strategy for Robot-Assisted Reminiscence Therapy}
%
\author{Fengpei Yuan\inst{1}\orcidID{0000-0002-5653-5552} \and
Ran Zhang\inst{2}\orcidID{0000-0002-6196-2890} \and
Dania Bilal\inst{1}\orcidID{0000-0002-4485-5647} \and
Xiaopeng Zhao\inst{1}\orcidID{0000-0003-1207-5379}}
\authorrunning{F. Yuan \emph{et al.}}
%
\institute{University of Tennessee, Knoxville TN 37996, USA \\
\email{fyuan6@vols.utk.edu; dania@utk.edu; xzhao9@utk.edu}\\
\and
Miami University, Oxford OH 45056, USA\\
\email{zhangr43@miamioh.edu}}
\maketitle              
\begin{abstract}
In this paper, the robot-assisted Reminiscence Therapy (RT) is studied as a psychosocial intervention to persons with dementia (PwDs). We aim at a conversation strategy for the robot by reinforcement learning to stimulate the PwD to talk. Specifically, to characterize the stochastic reactions of a PwD to the robot's actions, a simulation model of a PwD is developed which features the transition probabilities among different PwD states consisting of the response relevance, emotion levels and confusion conditions. A Q-learning (QL) algorithm is then designed to achieve the best conversation strategy for the robot. The objective is to stimulate the PwD to talk as much as possible while keeping the PwD's states as positive as possible. In certain conditions, the achieved strategy gives the PwD choices to continue or change the topic, or stop the conversation, so that the PwD has a sense of control to mitigate the conversation stress. To achieve this, the standard QL algorithm is revised to deliberately integrate the impact of PwD's choices into the Q-value updates. Finally, the simulation results demonstrate the learning convergence and validate the efficacy of the achieved strategy. Tests show that the strategy is capable to duly adjust the difficulty level of prompt according to the PwD's states, take actions (e.g., repeat or explain the prompt, or comfort) to help the PwD out of bad states, and allow the PwD to control the conversation tendency when bad states continue.

\end{abstract}
\section{Introduction}
Worldwide, approximately $50$ million people lived with dementia in 2018 \cite{Patterson2018world}.
Reminiscence therapy (RT), the most popular therapeutic intervention for persons with dementia (PwDs), exploits the PwDs' early memories and experiences, usually with some memory triggers familiar to the PwDs (e.g., photographs and music), to evoke memory and stimulate conversation \cite{woods2018reminiscence}. 
It has been evidenced that RT has positive effects on PwDs' quality of life, cognition, communication, and mood \cite{woods2018reminiscence}.
While computer-based RTs, such as the InspireD Reminiscence App \cite{ryan2020there} and Memory Tracks App \cite{cunningham2019assessing}, have been developed to make RT more accessible to PwDs,
the interaction modality is limited to 2D visual signals or sounds, lacking non-verbal interactions, e.g., eye gazing, body movement and facial expression. Comparatively, a physically embodied social robot capable of providing non-verbal interactions, is believed to enable more intuitive, effective and engaging memory triggers during RT \cite{yuan2021systematic}, thus stimulating more memory recall and conversation. In addition, robot-assisted RT is a promising solution to cope with the increasing number of PwDs and relieve the stress from the caregivers due to the dead-set execution and indefatigable repeatability \cite{Alz2021}. 

To train a robot to automate RT for PwDs, many types of learning algorithms have been proposed, including supervised learning, unsupervised learning, and reinforcement learning (RL). 
Caros \emph{et al.} \cite{caros2020automatic} applied deep learning technique to develop a smartphone-based conversational agent which automated RT by showing a picture, asking questions about the pictures, and giving comments on users' answers.
However, PwDs may have different dementia degrees, and an individual PwD may show time-varying behaviors, emotions, personalities, and cognitive capabilities \cite{cerejeira2012behavioral,kobayashi2019effects}. 
It is very challenging for supervised or unsupervised learning to achieve a learning agent with sufficient adaptivity to different individual PwDs. 
We herein target the robot training using RL, which allows the robot to constantly learn from interacting with the PwD and end up with an optimal conversation strategy for the target PwD \cite{hemminahaus2017towards}.
There are several existing works that investigated PwD-robot dialogue management using RL. For instance, Magyar \emph{et al.} \cite{magyar2019autonomous} employed Q-learning (QL) to learn a robotic conversation strategy to promote the PwD's response with considering the PwD's interested topics and emotions. Yuan \emph{et al.} \cite{yuan2021simulated} developed a robotic dialogue strategy via QL to handle the repetitive questioning behaviors from the PwDs. 
However, a pervasively applicable patient model is still lacking in the existing literature which can \emph{i}) integrate a comprehensive list of the major factors impacting the PwD's behaviors during RT, and \emph{ii}) accurately characterize the probabilistic transitions between the PwD's mental states under different robotic actions. Such a model will provide valuable guidance to more targetedly design the clinical experiments and collect the data, and serve as a customizable interface between the clinical data and the robotic RT strategy design. 

To this end, we aim to build a pervasive simulation model for PwDs to characterize their probabilistic behaviors during RT and develop a RL-based conversation strategy for robot-assisted RT. Specifically, our contributions are three-fold. 
Firstly, we design a parameterized pervasive PwD model which incorporates the PwD's response relevance, emotion levels and confusion conditions as the mental states, and depicts the probabilistic behaviors of PwDs during RT as probabilistic transitions between different mental states. Secondly, we define a Markov Decision Process (MDP) model for the robot-assisted RT and design a Q-learning (QL) algorithm to achieve the optimal conversation strategy for the robot. The strategy is sensitive to the PwD's mental states and promotes the PwD's talking by duly adjusting the difficulty level of the prompts, repeating or explaining the prompts to clear confusion, and comforting to help the PwD out of the bad moods. In case that bad moods continue, the strategy offers the PwD the initiative to continue or change the topic, or stop the conversation so that the stress of RT is mitigated. The impacts of the PwD's choices are also considered during the learning towards the optimal strategy. Finally, simulations are conducted to demonstrate the learning convergence and validate the efficacy of the achieved strategy in promoting conversation.

The remainder of the paper is organized as follows. Section \ref{Section:SimulationModel_PwD} describes the simulation model of PwD in the context of RT. Section \ref{Section:method} elaborates the design of the robotic conversation strategy, including the definition of MDP and the revised QL algorithm based on the proposed PwD model. The experimental results are presented and discussed in Section \ref{Section:result_Discussion} with suggested future work.

\section{Simulation Model for a Person with Dementia (PwD)}\label{Section:SimulationModel_PwD}

In the robot-assisted RT, the robot provides memory triggers (e.g., photographs, music, or video clips) and stimulates the PwD to talk about relevant past memory and experiences. During the conversation, the PwD with limited cognitive capacity may provide relevant, irrelevant, or even no response to the robot. In addition, the PwD may show different emotions, such as joy and discomfort, as a reaction to different memory triggers and the robot's actions. Moreover, the PwD may become confused about a question or memory trigger provided by the robot. Thus, in the context of robot-assisted RT, we represent the current state of a PwD by their response relevance, emotion levels, and confusion conditions. In this model, a PwD's response relevance can be relevant response (\textit{RR}), irrelevant response (\textit{IR}), or no response (\textit{NR}). A PwD's emotional level is categorized as negative (\textit{Neg}), neutral (\textit{Neu}), and positive (\textit{Pos}). A PwD's confusion condition is classified as confused (\textit{Yes}) and unconfused (\textit{No}).

Depending on the robot's actions during RT, the PwD's state may switch from one to another. We consider the following robot actions. On one hand, to stimulate the conversation during RT, the robot can provide appropriate prompts (e.g., questions) about the current memory trigger \cite{dijkstra2004conversational,small2012training}. The difficulty level of the prompt can be adjusted,  i.e., easy prompt ($a_1$, e.g., yes-no question), moderately difficult prompt ($a_2$), and difficult prompt ($a_3$, e.g., open-ended question).
On the other hand, when the PwD gets confused or in a negative emotion, conditions known as "harmful" or "bad" moments \cite{hofer2017reminiscence}, the robot may take actions to help the PwD out of these bad moments. Inspired by previous relevant studies \cite{hofer2017reminiscence,smith2011memory}, the robot will repeat ($a_4$) or explain ($a_5$) the prompt when the PwD feel confused \cite{dijkstra2004conversational,small2012training}, and comfort ($a_6$) the PwD to alleviate their fears or discomfort \cite{woods2012remcare} during RT.

The probabilistic behaviors of the PwD is modelled with the transition probabilities (See \href{https://drive.google.com/drive/folders/1FmhNsXJnG_WUUKtEpBBflig3ipks1qTb?usp=sharing}{Appendix I} or go to the next url: \url{https://drive.google.com/drive/folders/1FmhNsXJnG_WUUKtEpBBflig3ipks1qTb?usp=sharing}) among different PwD states given each robot action. Basically, as the difficulty level of the prompt from the robot increases (e.g., $a_3$ vs. $a_1$), the probabilities of the PwD responding relevantly and showing positive emotion will decrease, and the probabilities of getting confused will increase. When the PwD is in negative emotion or confused, the probabilities of relevant response will be smaller. If the PwD gets confused and the robot chooses to repeat or explain on the prompt, the probabilities of the PwD responding and showing non-negative emotion will increase, with the confusion condition possibly changed. If the robot chooses to comfort when the PwD is in bad moment, the probabilities of relevant response will be increased, with the emotion levels possibly changed better.

Moreover, our designed RT strategy will give the PwD initiative to control the conversation tendency when the bad moments continue. This will give the PwD a sense of control, thus mitigating the RT stress \cite{wilkins2017dementia}. If the PwD shows confusion or negative emotion continuously twice, the PwD will be provided with the choice of stopping the RT, continuing to talk about the current memory trigger, or changing to another memory trigger. If the PwD chooses to stop, the current RT session will terminate. If the PwD chooses to continue, the PwD's next state will remain unchanged. If the PwD chooses to change the memory trigger, the PwD's next state is considered to be no response (\emph{NR}), with neutral emotion (\emph{Neu}), and no confusion. We define the robot's action of providing choices as $a_7$.

Note that the state transition probabilities of different memory triggers are set to be identical in our simulations, but can be set different according to personal preferences \cite{gowans2004designing} in the future.

\section{Adaptive Robot-Assisted Reminiscence Therapy}\label{Section:method}
In this section, we apply the technique of reinforcement learning (RL) to learn a conversation strategy for the robot to deliver reminiscence therapy. The goal is to maintain the RT for a target number of conversation rounds, stimulate the PwD to express as much as possible, and keep the PwD's state as positive as possible. A revised Q-learning (QL) algorithm is used to achieve the best conversation strategy personalized to the PwD modelled in Section \ref{Section:SimulationModel_PwD}.

\subsection{Definition of Markov Decision Process}
In order to learn the optimal policy, we firstly formulate the problem of robot-assisted RT for PwD as the following MDP model \cite{sutton2018reinforcement}:

\textbf{State Space $S$.} 
A state $s$ in this problem is defined as the collection of the PwD's response relevance to the prompt from the robot, the emotion level, and the confusion condition. Based on the designed simulation model of PwD in Section \ref{Section:SimulationModel_PwD}, the state space $S$ has a cardinality of $3\times3\times2=18$.

\textbf{Action Space $A$.}
During RT, there are seven actions possibly taken by the robot, i.e., providing easy prompt ($a_1$), providing moderately difficult prompt ($a_2$), providing difficult prompt ($a_3$), repeating the prompt ($a_4$), briefly explaining the prompt ($a_5$), comforting the PwD ($a_6$), and giving the PwD choices ($a_7$).
Note that even the PwD responds to a prompt incorrectly, the robot will NOT correct the PwD (RT is not aimed to correct PwDs). However, the response relevance will be considered by the RL agent in the reward function.
Moreover, as mentioned previously, the robot will take action $a_7$ as long as the PwD shows confusion or negative emotion twice in a row. In other words, the condition of taking action $a_7$ is determinant, therefore, the actual action space for the RL only includes actions $a_1-a_6$. Although the Q values of taking action $a_7$ is not learned during the training, the impacts of taking $a_7$ is deliberately integrated into the Q-value update of other actions, which will be detailed in section \ref{subsec.train}.

\textbf{Reward Function $R$.}
The design of the reward function aligns with the objectives of the robot-assisted RT, i.e., stimulating the PwD to talk while keeping the PwD in a generally positive mood.
Thus, the reward function is a function of the PwD's response relevance, emotion level and confusion condition. Specifically, the robot should always try to prevent the PwD from getting trapped in the bad moments, i.e., being in negative mood or confused, as bad moments will hamper the conversation and lead to a higher chance of terminating the current session. Accordingly, the reward component of PwD's emotion level being negative, neutral, and positive is set to $-3$, $+1$, and $+2$, respectively. The reward component of the confusion condition is set to $-2.5$ and $+2$, respectively, for being confused and unconfused. As to the difficulty level of the prompt, it should be properly adjusted according to the PwD's cognitive capability and mental state so that the PwD is more engaged and interested, thus stimulating their memory and conversation to the most extent\cite{tapus2009use}. In other words, optimal tradeoff needs to be learnt between taking an easy prompt for higher chance of being in positive state and taking a more difficult prompt (e.g., an open question) to encourage the PwD to talk more. Correspondingly, we provide two reward settings (as listed in Table~\ref{tab:RewardFunct}) for prompts as a function of the difficulty level and resultant response relevance for later experimental study.
\begin{table}
\centering
\caption{Reward components w.r.t. response relevance in reward function $R_1$ and  $R_2$.}\label{tab:RewardFunct}
\begin{tabular}{|l|l|l|l|}
\hline
$R_1 / R_2$ & No response (NR) &  Irrelevant response (IR) & Relevant response (RR)\\
\hline
$a_2$ &  $-2 / -2$ & $0.75 / 0.75$ & $2/2$ \\
$a_3$ &  $-2/-2$ & $1.75 / 3$ & $3 / 10$ \\
$a_i$, $i\neq2,3$ & $-2/-2$ & $0.3/0.3$ & $0.75/0.75$ \\
\hline
\end{tabular}
\end{table}

\subsection{Learning Algorithm Design and Training}\label{subsec.train}
Although the RL agent only learns the optimal policy for taking actions $a_1-a_6$, the previously taken actions have decisive impact on the probability of taking action $a_7$. For example, if the robot always takes action $a_3$ (providing difficult prompt) and ignores PwD's bad moments, there will be a very high chance of the PWD choosing to stop. 
Therefore, we revise the standard QL algorithm and deliberately integrate the negative impact of taking $a_7$ in the Q-value updates of other actions to avoid over-aggressive policies, as summarized in Algorithm 1.
\begin{algorithm}
\caption{Revised Q-learning for robot-assisted RT}
\begin{algorithmic}[1]
\State Initialize $Q(s,a) \gets 0$, for all $s \in S$, $a \in A$
\Function{Loop for each episode}{}
\State Initialize $s_0 = [NR, Neu, NO]$, $Done = False$ 
\While{not $Done$}
\State Choose $a_t$ from $s_t$ using policy derived from $Q$ ($\epsilon$-greedy)
\State Take action $a_t$, observe $r_t$, $s_{t+1}$
\If{$a == a_6$}
  \State $Q(s_{t-1},a_{t-1}) \gets Q(s_{t-1},a_{t-1}) + \alpha [r_t + \gamma\max_{a}Q(s_{t+1},a) - Q(s_{t-1},a_{t-1})] $
\Else
  \State $Q(s_t,a_t) \gets Q(s_t,a_t) + \alpha [r_t + \gamma\max_{a}Q(s_{t+1},a_t) - Q(s_t,a_t)] $
\EndIf
\State $s_{t-1}, a_{t-1}, s_t \gets s_t, a_t, s_{t+1}$
\EndWhile
\EndFunction
\end{algorithmic}
\end{algorithm}

The RL agent is trained for $1500$ epochs, each with $30$ episodes. The learning rate and discount factor are set to be $0.05$ and $0.95$, respectively. At the beginning of each episode, the environment is reset to an initial state, [\textit{NR}, \textit{Neu}, \textit{No}]. In each iteration, the $\epsilon$-greedy approach ($\epsilon=0.1$) is used to select actions. An episode is terminated if the PwD chooses to stop, the maximum $50$ rounds are reached, or the number of memory triggers having been discussed reaches $15$.

\subsection{Evaluation Metrics}
To evaluate the performance, we compare the average return per epoch obtained by the revised QL (denoted as \textit{$\epsilon$-greedy QL}) to that obtained by a random policy (denoted as \textit{Random action}). Also, we extract the temporal policy $\pi'$ suggested at the $10$th Episode of each epoch, apply it to run $40$ experiments, and calculate the average return (denoted as \textit{Greedy QL}).
Moreover, we monitor the averaged sum of Q-table per epoch (i.e., Q-value sum) as well as its relative change (i.e., Q-value update) to evaluate the convergence performance.
Additionally, all the optimal policies suggested in the last $600$ episodes are recorded. We use the top five policies mostly suggested to run $1000$ experiments and choose the policy that obtains the maximum return as the final policy, denoted as $\pi_*'$.  Finally, we conduct 20 experiments with $\pi_*'$ and observe the dynamics of state-action transition in each experiment. 

\section{Results and Discussion}\label{Section:result_Discussion}
\begin{figure}
\includegraphics[width=\textwidth]{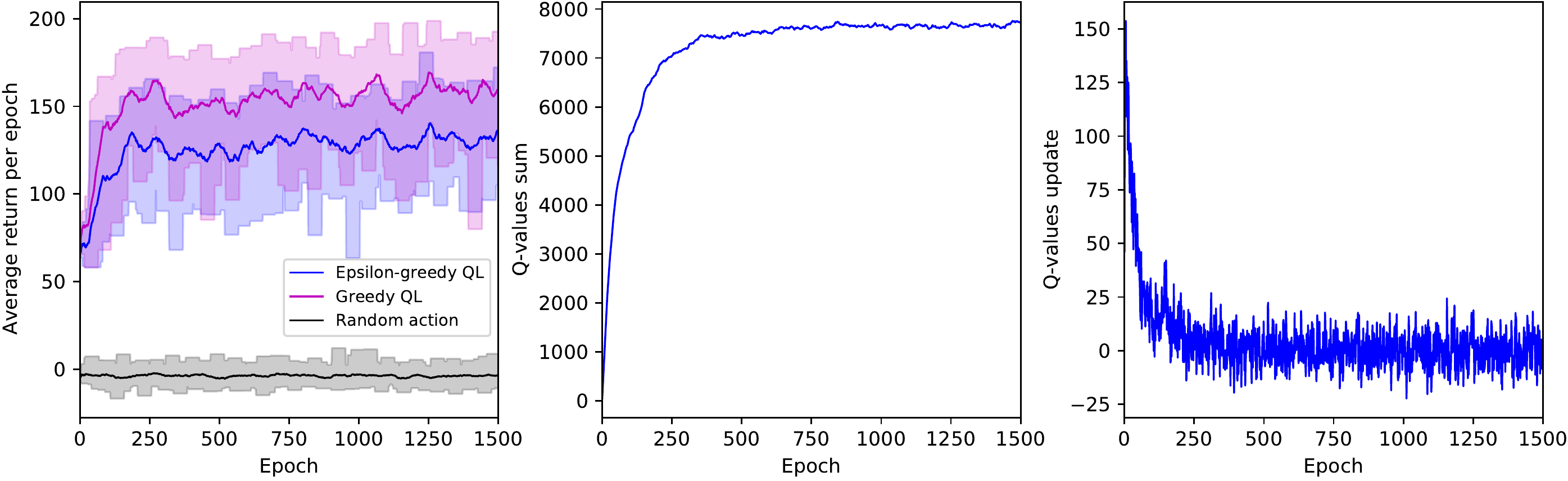}
\caption{The learning process by $\epsilon$-greedy Q-learning with reward function $R_1$.} \label{fig:Q_RewardFunct6}
\end{figure}

The learning process of our revised QL (i.e., $\epsilon$-greedy QL) with reward function $R_1$ is shown in Fig.~\ref{fig:Q_RewardFunct6}. As shown on the left of Fig.~\ref{fig:Q_RewardFunct6}, the average return per epoch obtained by $\epsilon$-greedy QL (blue curve) was much greater than the random action selection policy (black curve), which validated the efficacy of applying the RL approach for the robot to automate RT. The average return per epoch obtained by greedy QL was greater than $\epsilon$-greedy QL. This makes sense because the greedy QL always took optimal policy due to greedy action selection provided the achieved strategy is optimal, while the $\epsilon$-greedy QL selected random action for exploration.
With respect to the convergence, the curve of average return per epoch (blue curve in left figure of Fig.~\ref{fig:Q_RewardFunct6}) indicated the RL agent was able to converge within $200$ epochs, whereas the Q-values sum and Q-values update (the middle and right figure in Fig.~\ref{fig:Q_RewardFunct6}) converged within in $800$ epochs. Additionally, we observe that the optimal policy suggested by the RL agent in the last $600$ episodes is still changeable, which might be due to the design of reward function and the model of simulated PwD. In Table~\ref{tab:dynamicInExperimentDetail}, we listed the dynamics (e.g., state-action transition) of one experiment using the most nearly optimal action policy, $\pi_*'$, learned by Q-learning with reward $R_1$.
\begin{table}[p]
\caption{The PwD-robot interaction during a reminiscence therapy.}\label{tab:dynamicInExperimentDetail}
\centering
\begin{tabular}{|l|c|c||l|c|c|}
\hline
\makecell[c]{Step\\$0-25$} & \makecell[c]{State} &  \makecell[c]{Action\\(PwD's choice)} & \makecell[c]{Step\\$26-50$} & \makecell[c]{State} &  \makecell[c]{Action\\(PwD's choice)}\\
\hline
$0$ & $[0, 0, 0]$ & $0$ & $26$ & $[2, 1, 0]$ & $1$	\\
$1$ & $[0, 0, 0]$ & $0$	&$27$ & $[2, 1, 0]$ & $1$ \\
$2$ & $[0, 1, 0]$ & $0$	&$28$ & $[2, 1, 0]$ & $1$	\\
$3$ & $[0, 1, 1]$ & $3$	&$29$ & $[2, 1, 0]$ & $1$	\\
$4$ & $[2, 1, 1]$ & $6\rightarrow$(Change picture) & $30$ & $[2, 1, 0]$ & $1$\\
$5$ & $[0, 0, 0]$ & $0$	&$31$ & $[2, 1, 0]$ & $1$\\
$6$ & $[0, 1, 1]$ & $3$	& $32$ & $[2, 1, 0]$ & $1$	\\
$7$ & $[0, 1, 0]$ & $0$	&$33$ & $[1, 1, 0]$ & $0$\\
$8$ & $[1, 0, 0]$ & $0$	&$34$ & $[1, 1, 0]$ & $0$	\\
$9$ & $[2, 0, 0]$ & $0$	&$35$ & $[1, 1, 0]$ & $0$	\\
$10$ & $[2, 0, 0]$ & $0$ &$36$ & $[0, 0, 0]$ & $0$\\
$11$ & $[2, 0, 0]$ & $0$ &$37$ & $[1, 0, 1]$ & $3$ \\
$12$ & $[2, 0, 0]$ & $0$ &$38$ & $[1, 0, 0]$ & $0$	\\
$13$ & $[2, 0, 0]$ & $0$ &$39$ & $[0, 0, 1]$ & $4$ \\
$14$ & $[2, 0, 0]$ & $0$ &$40$ & $[0, 1, 0]$ & $0$ \\
$15$ & $[2, 0, 0]$ & $0$ &$41$ & $[0, 0, 0]$ & $0$ \\
$16$ & $[2, 0, 0]$ & $0$ &$42$ & $[2, 0, 0]$ & $0$ \\
$17$ & $[2, 0, 0]$ & $0$ &$43$ & $[2, 1, 0]$ & $1$ \\
$18$ & $[2, 0, 0]$ & $0$ &$44$ & $[2, 1, 0]$ & $1$ \\
$19$ & $[2, 0, 0]$ & $0$ &$45$ & $[2, 1, 0]$ & $1$ \\
$20$ & $[1, 0, 0]$ & $0$ &$46$ & $[2, 0, 0]$ & $0$ \\
$21$ & $[1, 0, 0]$ & $0$ &$47$ & $[2, 0, 0]$ & $0$ \\
$22$ & $[1, -1, 0]$ & $5$ &$48$ & $[2, 0, 0]$ & $0$ \\
$23$ & $[1, 0, 0]$ & $0$ &$49$ & $[2, 0, 0]$ & $0$ \\
$24$ & $[1, 1, 0]$ & $0$ &$50$ & $[2, 0, 0]$ & $0$ \\
$25$ & $[0, 1, 0]$ & $0$ & & & \\
\hline
\end{tabular}
\end{table}

\begin{figure}[p]
\includegraphics[width=\textwidth]{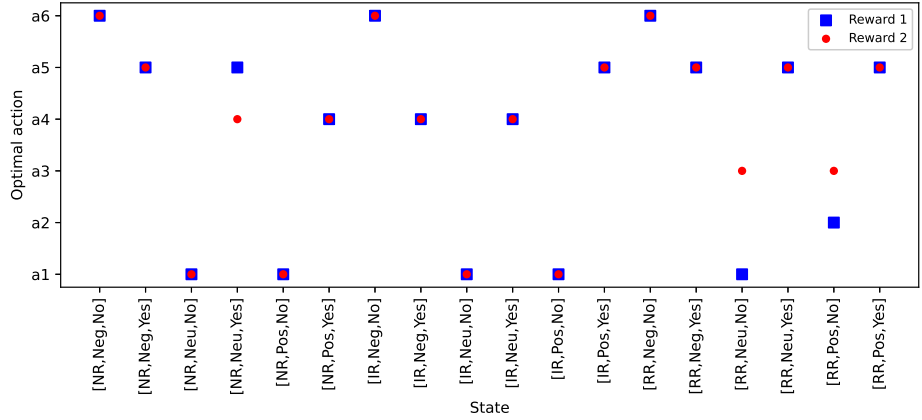}
\caption{The most nearly optimal policy $\pi_*'$ learned by $\epsilon$-greedy QL.} \label{fig:diff_twoRewardFunct}
\end{figure}

Compared to reward function $R_1$, Q-learning with reward function $R_2$ showed similar performance, i.e., curve of average return per epoch, converging Q-values sum, and changeable optimal policy during the last $600$ episodes. We present the most nearly optimal policy learned by QL using the two types of reward function, $R_1$ (the blue square) and $R_2$ (the red circle) in Fig.~\ref{fig:diff_twoRewardFunct}. \textcolor{blue}{note: 0,1,2}
The scatter plots demonstrates that the robot is able to comfort the PwD when they feel negative emotion. For example, the optimal actions suggested for state $s$ = [\textit{NR}, \textit{Neg},\textit{No}] with $R_1$ and $R_2$ were both $a_6$, comforting. When the PwD feel confused (e.g., $s$ = [\textit{NR}, \textit{Pos},\textit{Yes}]), the RL agent with $R_1$ and $R_2$ both suggested to take action, $a_4$, repeating the prompt to the PwD.
There were $3$ states, $s=$[\textit{NR}, \textit{Neu},\textit{Yes}], [\textit{RR}, \textit{Neu}, \textit{No}], and [\textit{RR}, \textit{Pos}, \textit{No}], where the RL agent suggested different actions regarding $R_1$ and $R_2$. The RL agent with $R_1$ suggested to take action $a_0$ and $a_1$ when the PwD in a state of $s$=[\textit{RR}, \textit{Neu}, \textit{No}] and [\textit{NR}, \textit{Pos}, \textit{No}], respectively. Comparatively, the RL agent with $R_2$ would take action $a_2$ when the PwD in state $s$=[\textit{RR}, \textit{Neu}, \textit{No}] or [\textit{NR}, \textit{Pos}, \textit{No}].
Such difference make sense because the two types of reward function $R_1$ and $R_2$ (in Table.~\ref{tab:RewardFunct}) indicated how much we value PwD's response relevance and the level of PwD's conversation being stimulated.
The reward function $R_2$ was more aggressive compared to $R_1$, that is, the memory and conversation being stimulated was much more valued by $R_2$ than the condition of emotion and confusion. On the other hand, this scatter plot also indicates that our RL approach was able to learn to adjust the difficulty level of prompt adaptive to PwD's conditions.

In this paper, we employed a revised QL to learn a conversation strategy for the robot to stimulate a PwD to talk as much as possible while keeping the PwD in a generally positive mood during RT. The PwD was modelled as the transition probabilities among different conditions consisting of the response relevance, emotion levels and confusion status. Our experimental results showed that the strategy learned by QL was capable to adjust the difficulty level of prompt (e.g., yes-no vs. open-ended question) according to the PwD's states, take actions such as repeating/explaining the prompt or comforting to help the PwD out of bad moments \cite{smith2011memory,hofer2017reminiscence}, and allow the PwD to mitigate potential conversation stress during RT.
To the best of our knowledge, this is the first time for technology-enabled RT to learn adaptive strategy, while taking into consideration of complicated PwDs' mental states and communication strategies suggested in the traditional healthcare field. This might offer a promising solution for automatic, person-centered RT for PwD living alone. 


However, there are still some limitations in this study.
The patient model, the matrix of state-action transition probabilities, was created based on previous qualitative studies. As we discussed earlier, the nature underlying our PwD's model might result in the optimal policy was still changing during the last $600$ episodes.
For better learning of RL for RT as well as the real-world application of robot-assisted RT, a patient model based on real-world data should be developed, which is our next step. Additionally, we designed two types of reward function to test the feasibility of RL approach. However, the design of reward function might be associated with a PwD's own personality and needs (e.g., psychological needs vs cognitive stimulation). From this perspective, in future work, we will closely collaborate with professional facilitators in this field and PwDs to adjust the reward function, to ensure an effective, person-centered RT using RL.

\bibliographystyle{splncs04}

\bibliography{Reference}

\begin{thebibliography}{10}
\providecommand{\url}[1]{\texttt{#1}}
\providecommand{\urlprefix}{URL }

\bibitem{Patterson2018world}
Patterson, C.: World alzheimer report 2018. Alzheimer’s Disease International
  (ADI), London  (September 2018)

\bibitem{woods2018reminiscence}
Woods, B., O'Philbin, L., Farrell, E.M., Spector, A.E., Orrell, M.:
  Reminiscence therapy for dementia. Cochrane database of systematic reviews
  (3) (2018)

\bibitem{ryan2020there}
Ryan, A.A., McCauley, C.O., Laird, E.A., Gibson, A., Mulvenna, M.D., Bond, R.,
  Bunting, B., Curran, K., Ferry, F.: ‘there is still so much inside’: The
  impact of personalised reminiscence, facilitated by a tablet device, on
  people living with mild to moderate dementia and their family carers.
  Dementia  19(4),  1131--1150 (2020)

\bibitem{cunningham2019assessing}
Cunningham, S., Brill, M., Whalley, J.H., Read, R., Anderson, G., Edwards, S.,
  Picking, R.: Assessing wellbeing in people living with dementia using
  reminiscence music with a mobile app (memory tracks): a mixed methods cohort
  study. Journal of healthcare engineering  2019 (2019)

\bibitem{yuan2021systematic}
Yuan, F., Klavon, E., Liu, Z., Lopez, R.P., Zhao, X.: A systematic review of
  robotic rehabilitation for cognitive training. Frontiers in Robotics and AI
  8,  105 (2021)

\bibitem{Alz2021}
2021 alzheimer's disease facts and figures. Alzheimer's \& dementia: the
  journal of the Alzheimer's Association  17(3),  327–406 (2021)

\bibitem{caros2020automatic}
Caros, M., Garolera, M., Radeva, P., Giro-i Nieto, X.: Automatic reminiscence
  therapy for dementia. In: Proceedings of the 2020 International Conference on
  Multimedia Retrieval. pp. 383--387 (2020)

\bibitem{cerejeira2012behavioral}
Cerejeira, J., Lagarto, L., Mukaetova-Ladinska, E.: Behavioral and
  psychological symptoms of dementia. Frontiers in neurology  3, ~73 (2012)

\bibitem{kobayashi2019effects}
Kobayashi, M., Kosugi, A., Takagi, H., Nemoto, M., Nemoto, K., Arai, T.,
  Yamada, Y.: Effects of age-related cognitive decline on elderly user
  interactions with voice-based dialogue systems. In: IFIP Conference on
  Human-Computer Interaction. pp. 53--74. Springer (2019)

\bibitem{hemminahaus2017towards}
Hemminahaus, J., Kopp, S.: Towards adaptive social behavior generation for
  assistive robots using reinforcement learning. In: 2017 12th ACM/IEEE
  International Conference on Human-Robot Interaction (HRI. pp. 332--340. IEEE
  (2017)

\bibitem{magyar2019autonomous}
Magyar, J., Kobayashi, M., Nishio, S., Sin{\v{c}}{\'a}k, P., Ishiguro, H.:
  Autonomous robotic dialogue system with reinforcement learning for elderlies
  with dementia. In: 2019 IEEE International Conference on Systems, Man and
  Cybernetics (SMC). pp. 3416--3421. IEEE (2019)

\bibitem{yuan2021simulated}
Yuan, F., Sadovnik, A., Zhang, R., Casenhiser, D., Paek, E.J., Yoon, S.O.,
  Zhao, X.: A simulated experiment to explore robotic dialogue strategies for
  people with dementia. arXiv preprint arXiv:2104.08940  (2021)

\bibitem{dijkstra2004conversational}
Dijkstra, K., Bourgeois, M.S., Allen, R.S., Burgio, L.D.: Conversational
  coherence: Discourse analysis of older adults with and without dementia.
  Journal of Neurolinguistics  17(4),  263--283 (2004)

\bibitem{small2012training}
Small, J., Ann~Perry, J.: Training family care partners to communicate
  effectively with persons with alzheimer's disease: The traced program.
  Canadian Journal of Speech-Language Pathology \& Audiology  36(4) (2012)

\bibitem{hofer2017reminiscence}
Hofer, J., Busch, H., {\v{S}}olcov{\'a}, I.P., Tavel, P.: When reminiscence is
  harmful: The relationship between self-negative reminiscence functions, need
  satisfaction, and depressive symptoms among elderly people from cameroon, the
  czech republic, and germany. Journal of Happiness Studies  18(2),  389--407
  (2017)

\bibitem{smith2011memory}
Smith, E.R., Broughton, M., Baker, R., Pachana, N.A., Angwin, A.J., Humphreys,
  M.S., Mitchell, L., Byrne, G.J., Copland, D.A., Gallois, C., et~al.: Memory
  and communication support in dementia: research-based strategies for
  caregivers. International Psychogeriatrics  23(2),  256 (2011)

\bibitem{woods2012remcare}
Woods, R.T., Bruce, E., Edwards, R., Elvish, R., Hoare, Z., Hounsome, B.,
  Keady, J., Moniz-Cook, E., Orgeta, V., Orrell, M., et~al.: Remcare:
  reminiscence groups for people with dementia and their family
  caregivers-effectiveness and cost-effectiveness pragmatic multicentre
  randomised trial. Health Technology Assessment  16(48) (2012)

\bibitem{wilkins2017dementia}
Wilkins, J.M.: Dementia, decision making, and quality of life. AMA J Ethics
  19(7),  637--639 (2017)

\bibitem{gowans2004designing}
Gowans, G., Campbell, J., Alm, N., Dye, R., Astell, A., Ellis, M.: Designing a
  multimedia conversation aid for reminiscence therapy in dementia care
  environments. In: CHI'04 Extended Abstracts on Human Factors in Computing
  Systems. pp. 825--836 (2004)

\bibitem{sutton2018reinforcement}
Sutton, R.S., Barto, A.G.: Reinforcement learning: An introduction. MIT press
  (2018)

\bibitem{tapus2009use}
Tapus, A., Tapus, C., Mataric, M.J.: The use of socially assistive robots in
  the design of intelligent cognitive therapies for people with dementia. In:
  2009 IEEE international conference on rehabilitation robotics. pp. 924--929.
  IEEE (2009)

\end{thebibliography}





\end{document}